\documentclass[lettersize,journal]{IEEEtran}
\usepackage{amsmath,amsfonts}
\usepackage{algorithmic}
\usepackage{algorithm}
\usepackage{array}
\usepackage[caption=false,font=normalsize,labelfont=sf,textfont=sf]{subfig}
\usepackage{textcomp}
\usepackage{stfloats}
\usepackage{url}
\usepackage{verbatim}
\usepackage{graphicx}

\usepackage{cite}

\usepackage{amsmath}
\usepackage{amssymb}
\usepackage{booktabs}
\usepackage{multirow}
\usepackage{tabularray}

\usepackage{booktabs}
\usepackage{multirow}
\usepackage{graphicx}
\usepackage[table,xcdraw]{xcolor}
\usepackage[normalem]{ulem}
\usepackage{subcaption}
\useunder{\uline}{\ul}{}

\begin{document}

\title{SpirDet: Towards Efficient, Accurate and Lightweight Infrared Small Target Detector}

\author{Qianchen Mao, Qiang Li, Bingshu Wang*, Yongjun Zhang, Tao Dai, C.L. Philip Chen, \IEEEmembership{Fellow, IEEE}
\thanks{This work is supported by the National Natural Science Foundation of China, Youth Fund, under number 62102318, and the Basic Research Programs of Taicang, 2023, under number TC2023JC23. This work is also funded in part by the National Natural Science Foundation of China grant under number 92267203, in part by the Science and Technology Major Project of Guangzhou under number 202007030006, and in part by the Program for Guangdong Introducing Innovative and Entrepreneurial Teams (2019ZT08X214). (\it{Corresponding author: Bingshu Wang.})}
\thanks{Qianchen Mao is with the School of Software, Taicang Campus, Northwestern Polytechnical University, Suzhou 215400, China (e-mail: mqc@mail.nwpu.edu.cn).}
\thanks{Qiang Li is with the School of Software, Taicang Campus, Northwestern Polytechnical University, Suzhou 215400, China (e-mail: lq2023@mail.nwpu.edu.cn).}
\thanks{Bingshu Wang is with the School of Software, Northwestern Polytechnical University,
Xi’an 710129, China, and National Engineering Laboratory for Big Data System Computing Technology, Shenzhen University, Shenzhen 518060, China. (e-mail: wangbingshu@nwpu.edu.cn).}
\thanks{Yongjun Zhang is with the State Key Laboratory of Public Big Data, College of Computer Science and Technology, Guizhou University, Guiyang
550025, Guiyang, China. (e-mail: zyj6667@126.com).}
\thanks{
Tao Dai is with the College of Computer Science and Software Engineering, Shenzhen University, Shenzhen 518060, China (e-mail: daitao.edu@gmail.com).}
\thanks{C. L. P. Chen is with the School of Computer Science and Engineering, South China University of Technology and Pazhou Lab, Guangzhou 510641, 510335, China (e-mail: philip.chen@ieee.org).}}

\markboth{Journal of \LaTeX\ Class Files,~Vol.~14, No.~8, August~2021}%
{Shell \MakeLowercase{\textit{et al.}}: A Sample Article Using IEEEtran.cls for IEEE Journals}



\maketitle

\begin{abstract}
   In recent years, the detection of infrared small targets using deep learning methods has garnered substantial attention due to notable advancements. To improve the detection capability of small targets, these methods commonly maintain a pathway that preserves high-resolution features of sparse and tiny targets. However, it can result in redundant and expensive computations. To tackle this challenge, we propose SpirDet, a novel approach for efficient detection of infrared small targets. Specifically, to cope with the computational redundancy issue, we employ a new dual-branch sparse decoder to restore the feature map. Firstly, the fast branch directly predicts a sparse map indicating potential small target locations (occupying only 0.5\% area of the map). Secondly, the slow branch conducts fine-grained adjustments at the positions indicated by the sparse map. Additionally, we design an lightweight DO-RepEncoder based on reparameterization with the Downsampling Orthogonality, which can effectively reduce memory consumption and inference latency. Extensive experiments show that the proposed SpirDet significantly outperforms state-of-the-art models while achieving faster inference speed and fewer parameters. For example, on the IRSTD-1K dataset, SpirDet improves $MIoU$ by 4.7 and has a $7\times$ $FPS$ acceleration compared to the previous state-of-the-art model. The code will be open to the public. 
\end{abstract}


\section{Introduction}
\label{sec:intro}

\begin{figure}[h]
\centering
\centerline{\includegraphics[width=0.4\textwidth,trim=15 0 0 10,clip]{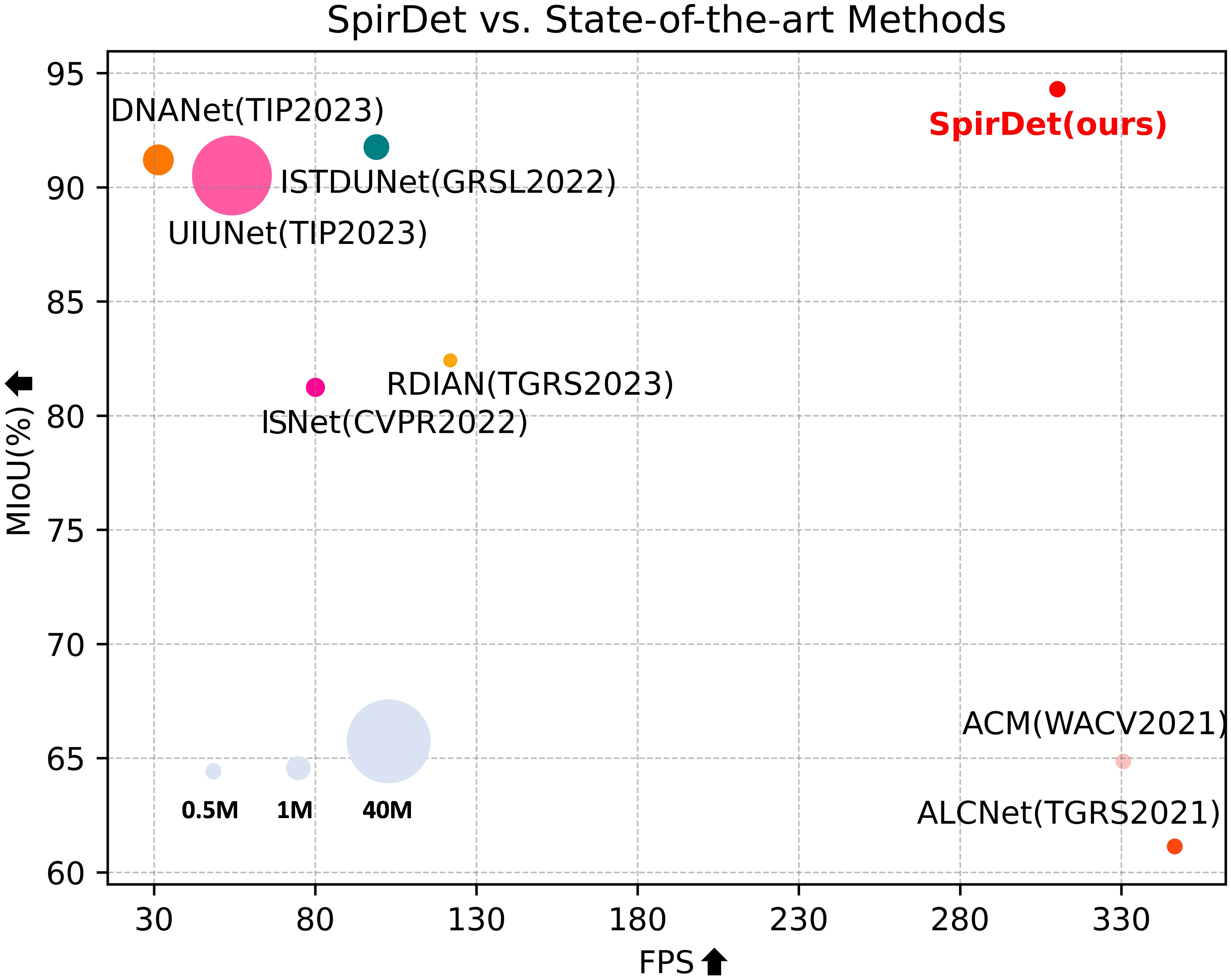}}
\caption{The $MIoU$, $FPS$ versus model size on NUDT-SIRST\cite{DNANET} test set. Our SpirDet are marked as red point. As can be seen from the comparison, our SpirDet achieves a better $FPS$/$MIoU$. All the frame per second($FPS$) were measured on a single NVIDIA RTX 3090, with an input size of $256 \times 256$.}\label{fig:score1}
\end{figure}
Infrared small target detection holds significant potentials for various applications\cite{rawat2020review,2021review,2022survey}, such as maritime rescue\cite{zhang2022rkformer,zhang2022exploring} and urban security\cite{urban}. In these situations, it is vital for the detectors to identify targets while ensuring both high accuracy and enhanced inference speed\cite{wang2019novel}. Hence, investigating a high-performance and efficient infrared small target detector is a prominent area of focus.

Infrared sensors capture solely thermal radiation signals, leading to images lacking of details\cite{2023review}. It is necessary for an infrared small target detector to discern targets from a significant distance. As a result, these targets typically occupy the image only a sparse number of pixels, ranging from a few to several tens.  This  forms two characteristics of infrared small targets: 1) small and weak\cite{yu2022infrared,zuo2022affpn}, due to their minute size, discernible texture, shape, and other target information are significantly diminished. 2) low signal-to-noise ratio\cite{ma2022infrared,zhang2022research,yang2022infrared}, the abundance of noise interference within infrared images can lead to false identification of background interference as targets.

Early detection of infrared small targets are model-based methods\cite{LCM,PFT,LRSR}, leveraging human prior knowledge for detection and offering commendable real-time performance. Nonetheless, these methods are significantly impeded by the hand-crafted features, resulting in poor generalization, low accuracy and high false detection rates. Recent advancements redefine infrared small target detection as a semantic segmentation task\cite{iaanet,eaaunet,MSvsFA}, utilizing architectures such as U-net series\cite{ronneberger2015u}. Given that the targets occupy only a sparse subset of the image's pixels, there is a potential risk of spatial information loss during the down-sampling phase within the encoder. To enhance the performance of small target detection, some methods  \cite{DNANET,istdunet,uiunet} maintain a high-resolution feature map pathway between the encoder and decoder. However, as the feature map size increases, the computational cost grows quadratically.



The motivation of this paper is to focus computations on the potential positions of small targets within high-resolution (HR) feature maps for tasks necessitating high resolution, such as edge detection and shape learning of small targets. To achieve high speed and superior performance, this paper introduces SpirDet, which capitalizes on several distinctive characteristics of infrared small targets.
 We propose the Dual-branch Sparse Decoder(DBSD). This decoder utilizes a fast branch to predict the coarse-grained positions of small targets on low-resolution feature maps directly, followed by a slow branch that employs sparse convolution to refine the positions within these coarse-grained regions on an HR level. This refinement compensates for performance degradation experienced when restoring low-resolution maps to their original scale. Furthermore, we design a reparameterized Encoder, termed DO-RepEncoder, which boasts a large model capacity while preserving a high inference speed.

As shown in Fig. \ref{fig:score1}, our SpirDet appreciably improves Mean Intersection over Union ($MIoU$) while markedly reducing inference latency. It surpasses existing methods for infrared small target detection in terms of metrics such as the false alarm rate ($F_a$), probability of detection ($P_d$), and Frames per Second($FPS$). 

Our contributions can be summarized as follows:

$\bullet$ We propose a high-performance and high-speed network architecture called SpirDet for infrared small target detection. It incorporates a reparameterization module into DO-RepEncoder to maintain model capacity while substantially accelerating model inference speed. 
 
$\bullet$ We present a Dual-branch Sparse Decoder (DBSD) including fast and slow branch. The fast branch produces a sparse map of potential target positions and guides the slow branch for high-resolution and fine-grained refinement. This can reduce computation costs of model largely. 

$\bullet$ Experiments conducted on the NUDT-SIRST\cite{DNANET} dataset demonstrate that the SpirDet exceeds the state-of-the-art (SOTA) by 2.09\% , accompanied by a threefold increase in $FPS$. Similarly, on the IRSTD-1K\cite{isnet} dataset, SpirDet attains SOTA results with a 4.74\% improvement, boasting an $FPS$ that is 7$\times$ higher than the previous SOTA.



\section{Related Work}
\label{sec:formatting}
\begin{figure*}[ht!]
\centering
\centerline{\includegraphics[width=\textwidth,trim=0 350 350 50,clip]{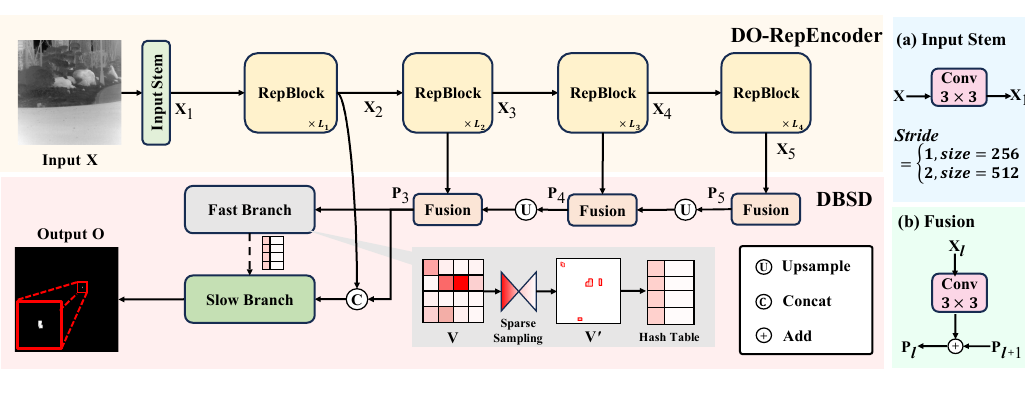}}
\caption{The overall architecture of the proposed SpirDet, which comprises DO-RepEncoder and Dual-branch Sparse Decoder. }
\label{fig:architecture}
\end{figure*}

\subsection{Infrared Small Target Detection}
The prevailing trends in Infrared Small Target Detection (IRSTD) coalesce into two principal categories: model-based methods and data-driven methods. Model-based methods\cite{random,mg} leverage artificial prior knowledge for detection, encompassing strategies that suppress backgrounds\cite{tophat,maxmedian}, construct models informed by the human visual system\cite{LCM,JLCM,MSLCM,tllcm,wslcm,mcm}, and employ optimization-based techniques\cite{ipi,nram}. Despite their ability to execute swiftly on edge devices, these methods frequently fall short in terms of generalization capability in complicated scenarios.

Conversely, the latter category interprets IRSTD as a semantic segmentation problem underpinned by deep learning\cite{sddnet,filterdet,oscar,Marine,peng2023dynamic}. Some approaches\cite{acm,ALCNet,bprnet,istdunet,uiunet} capture high-level semantic features via cross-level feature fusion to obtain high-resolution, detailed small target features. Another strain of methods concentrates on improving the learning of small target context information representation\cite{dim2clear}, generally employing Attention mechanisms\cite{DNANET,rdian,swin,CAFFNET,li2023transnet,courtnet}. In addition, certain methods direct the network's focus on the edges and shape information of infrared small targets, thereby enabling more accurate representation of the small target. 

 Data-driven methods have demonstrated superior results compared to model-based methods. However, these methods grapple with achieving a balance between speed and performance due to the conflict between high-resolution small target feature learning and low network inference latency. In this paper, we address this contradiction through the implementation of sparse convolution.

\subsection{Sparse Convolution Network}
Sparse convolution networks have recently been recognized as an effective technique for managing tasks predicated on sparse prior knowledge. For example, Yan \textit{et al.} \cite{second} utilized 3D sparse convolution in the 3D object detection with voxel representation, jump over non-object features and thereby increasing the speed of network inference. In the realm of target detection, Yang \textit{et al.} \cite{yang2022querydet} employed sparse convolution to pinpoint small targets within the high-resolution feature maps of the FPN\cite{lin2017feature}. Du \textit{et al.} \cite{du2023adaptive} trained masks using the Gumbel-Softmax\cite{verelst2020dynamic} trick, leading to the adaptive acceleration of heavy detection heads via sparse convolution. 

In this paper, sparse convolution network is incorporated into IRSTD. By learning the sparse map of potential small target locations in the low-resolution feature map of the Decoder, we guide sparse convolution to specifically fine-tune the high-resolution information of small targets.

\section{Method}

\subsection{Overall Architecture}
\label{subsec:Overall Architecture}
The overall architecture of the proposed SpirDet is illustrated in Fig. \ref{fig:architecture}. It comprises DBSD and DO-RepEncoder. The network takes an infrared image $\mathbf{X} \in \mathbb{R}^{1 \times H \times W}$ as input and initially reformats it to the required input shape using the Input Stem, as depicted in Fig. \ref{fig:architecture} (a). The DO-RepEncoder, composed of four stages of RepBlocks, performs downsampling using $3 \times 3$ convolutions in the first block of RepBlock. Following this, the feature maps $\mathbf{P}_i$ at varying levels undergo an upsampling process and lightweight fusion operations to capture multi-scale and contextual information. As demonstrated in Fig. \ref{fig:architecture} (b), the lightweight fusion concatenates $\mathbf{P}_{l+1}$ with the shallower feature maps, succeeded by a $3 \times 3$ convolution, batch normalization (BN), and ReLU activation, before direct addition to $\mathbf{P}_l$. Finally, the DBSD processes the feature maps, incorporating a Fast-Slow Dual-branch, to output the final result $\mathbf{O}$. 


\subsection{Dual-branch Sparse Decoder}
\label{subsec:Dual-branch Sparse Decoder}
\begin{figure}[ht]
\centering
\centerline{\includegraphics[width=0.36\textwidth,trim=0 0 0 5,clip]{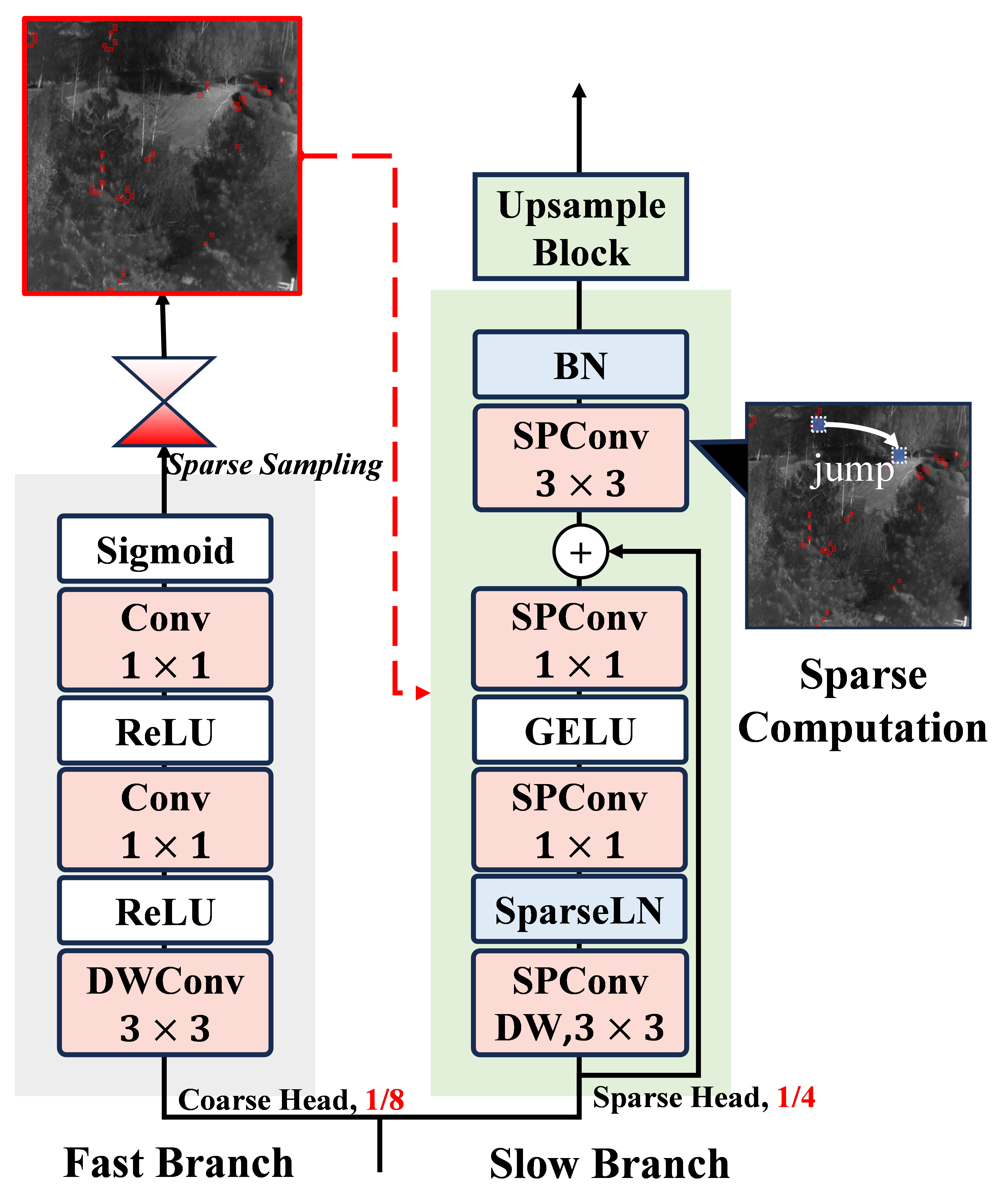}}
\caption{The structure diagram of the DBSD. The term `Sparse Computation' signifies that the sparse convolution operates exclusively at significant locations. \textbf{1/8} and \textbf{1/4} denote one-eighth and one-fourth of the original resolution, respectively. The red areas in the image represent potential small target locations. }
\label{fig:sparse}
\end{figure}

\begin{figure*}[ht]
\centering
\centerline{\includegraphics[width=0.8\textwidth,trim=0 370 350 60,clip]{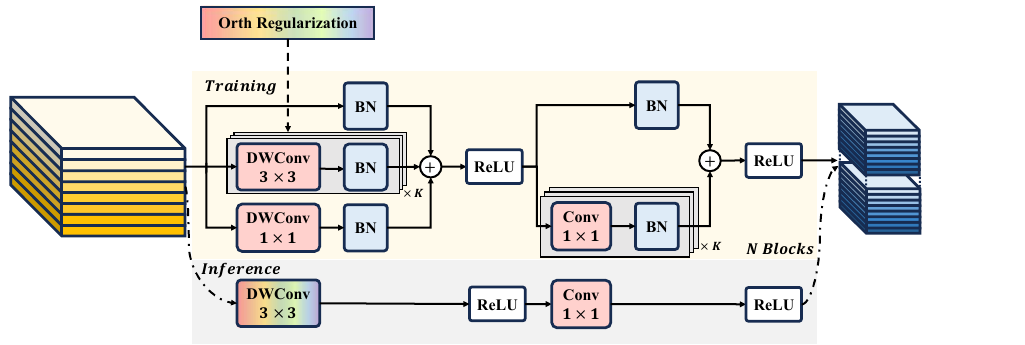}}
\caption{The framework of RepBlock. It is composed of \textit{N} blocks, each of which contributes to a multi-branch structure during the training phase and is reparameterized into a single-branch structure in the inference phase. }
\label{fig:RepEncoder}
\end{figure*}

To mitigate the potential loss of small target features when downsampling, prevalent architectures of infrared small target detectors typically sustain a high-resolution feature map pathway to secure clear target features. However, the distribution of infrared small targets within images is generally extremely sparse. An intuitive approach is to restrict computations exclusively to target-related features, automatically bypassing target-irrelevant features when convolution operates on the feature maps. The computation guided by this approach partially resolves the significant imbalance issue between foreground and background classes that is frequently encountered in infrared small target detection.

In this section, we introduce the Dual-branch Sparse Decoder (DBSD), which incorporates both the \textit{fast branch} and the \textit{slow branch}. The \textit{fast branch}, outfitted with the \textit{coarse head}, swiftly predicts the coarse-grained potential locations of small targets with minimal computational costs on lower-resolution feature maps. These potential locations are then subjected to \textit{sparse sampling} to yield a sparse binary mask with a low ratio (for instance, 0.5\% of the entire image). The sparse binary mask is further transformed into a Hash Table indicating the positions. This hash table directs the \textit{sparse head} of the \textit{slow branch} to execute concentrated computations on the potential locations of small targets on higher-resolution feature maps. Subsequently, the predicted feature map is returned to the original resolution via the \textit{upsample block}. The implementation specifics of DBSD are depicted in Fig. \ref{fig:sparse}.

The \textit{fast branch} is engineered to acquire the sparse mask. For instance, when given an original resolution input $\mathbf{X} \in \mathbb{R}^{ 1 \times H_l \times W_l}$, the \textit{coarse head} accepts a lower-resolution feature map $\mathbf{P_l} \in \mathbb{R}^{ C_l \times H_l \times W_l}$ from the $l$-th layer of the decoder, producing a probability map $\mathbf{V} \in \mathbb{R}^{1 \times H_l \times W_l}$ that signifies the likelihood of the grid $\mathbf{V}_{i,j}$ being associated with an infrared small object. Here, $C_l$, $H_l$, and $W_l$ denote the number of channels, height, and width of the $l$-th layer, respectively. To streamline the prediction task of the \textit{coarse head}, we employ the SoftIoU loss to supervise the output probability map $\mathbf{V}$ as:
\begin{equation}
    {\mathcal{L}}_{sparse}(\mathbf{V}, \mathbf{GT})=1- \mathit{SoftIoU} (\mathbf{V}, \mathit{Maxpool} (\mathbf{GT} )),
\end{equation}
where the coarse-grained ground truth (GT) is derived using the Maxpooling operator. Consequently, the \textit{coarse head} can confidently predict the probability map $\mathbf{V}$ of potential small object locations. The probability map $\mathbf{V}$ then undergoes $TOP-K$ filtering, retaining only the grids exhibiting the highest confidence to form a sparse binary map $\mathbf{V}^{'} \in \left\{0,\,1\right\}^{1\times H\times W}$, where a hyperparameter $\alpha$ (e.g., 0.5\%) determines the retention ratio of $\mathbf{V}$. This step is referred to as \textit{sparse sampling}:
\begin{equation}
\mathbf{V}^{'}_{i,j} = \begin{cases} 
1 & \text{if } \text{i, j} \in \text{TOP-K}(\mathbf{V}, \alpha\times H\times W) \\
0 & \text{otherwise}
\end{cases},
\label{eq:v'}
\end{equation}
where the $TOP-K$ returns the 2D coordinates of the filtered grid. 

The motivation for \textit{sparse sampling} is twofold: (1) managing the scale of the overall sparse region, which only has a noticeable acceleration effect when scale is small, and (2) utilizing $\alpha$ to control the range of the context region related to small targets. Because infrared small targets are exceedingly small, the context region genuinely associated with the target is also extremely sparse. Thus, by adjusting $\alpha$, the context region related to small targets can be compressed drastically. The performance metrics are substantially influenced by the hyperparameter $\alpha$, which will be discussed in the experimental section.

The \textit{slow branch} ingests a higher-resolution feature map, denoted as $\mathbf{P}_{l-1}$, and generates a probability map that matches the original resolution. Subsequently, $\mathbf{P}_{l-1}$ is dispatched to the \textit{sparse head}, an efficient component that capitalizes on sparse convolution. This type of convolution selectively operates on the foreground region using the sparse binary mask $\mathbf{V}^{'}$, as detailed in Eq. \ref{eq:v'}. Throughout the network inference process, convolution is strictly applied where the mask value is 1. This strategic utilization of sparsity significantly reduces computational requirements when processing high-resolution target-related features. Unlike traditional convolutions, which uniformly compute on each grid of the feature map, the \textit{sparse head}'s computation exhibits spatial variation. It directly processes information related to small targets, taking advantage of the region of interest (ROI) role played by the sparse map $\mathbf{V}^{'}$. This approach enables it to more effectively learn intricate details such as shape.  The \textit{sparse head} further reduces computation by utilizing efficient combinations of DWConv and PWConv, and subsequently integrates spatial and channel information using a $3 \times 3$ Conv. Ultimately, the feature map generated by the \textit{sparse head} is combined with a straightforward $1$-Conv+$1$-Bilinear operation to produce the output $\mathbf{O} \in \mathbb{R}^{B \times 1 \times H_l \times W_l}$.

\subsection{DO-RepEncoder}
\label{subsec:RepEncoder with Downsampling Orthogonality}


To reach a trade-off between model performance and inference speed, we introduce the reparameterization technique to satisfy model capacity and speed. Inspired by \cite{ding2021repvgg,vasu2023mobileone}, we design the DO-RepEncoder for feature extraction as shown in Fig. \ref{fig:RepEncoder}. The DO-RepEncoder consists of stages at different resolutions, with each Stage $i$ containing $L_i$ RepBlocks. Each RepBlock incorporates a Depthwise part and a Pointwise part, the former featuring a $3 \times 3$ convolution and the latter, a $1 \times 1$ convolution, both comprising a $K$ parallel convolutions branch, which is represented by:
\begin{equation}
\text{RepBlock}(\mathbf{X}_i) = \text{Pointwise}(\text{Depthwise}(\mathbf{X}_i)).
\end{equation}
During training, the Depthwise part caculated as below:
\begin{equation}
\begin{aligned}
\text{Depthwise}(\mathbf{X}_i) &=ReLU(BN(\mathbf{X}_i)\\&~~~~+\sum_{k}^{K} BN(3\times 3 DWConv_k(\mathbf{X}_i))\\&~~~~+BN(1\times 1 DWConv(\mathbf{X}_i))).
\end{aligned}
\end{equation}
During inference, it transitions to a single-branch structure, ensuring the model's speed:
\begin{equation}
\text{Depthwise}(\mathbf{X}_i) = ReLU(3\times 3 DWConv(\mathbf{X}_i)).
\end{equation}
The multi-branch design during training enhances the overall model capacity.

To prevent the diminution of small object features during the downsampling process, we propose a \textit{downsampling orthogonality} method by incorporating \textit{orthogonality regularization}\cite{biggan,brock2016neural} into the initial downsampling block of each stage, advocating for orthogonal weights among the K parallel convolutions:
\begin{equation}    {\mathcal{L}}_{orth}(\mathbf{W_{down}})=\left \| \mathbf{W_{down}}^T\mathbf{W_{down}}-I \right \|_{F}^{2}  ,\end{equation}
where $\mathbf{W_{down}} \in \mathbb{R}^{(out \times K) \times in \times 3 \times 3}$ is
\begin{equation}    \mathbf{W_{down}}=Concat(W_1,W_2,...,W_K).\end{equation}
Incorporating \textit{downsampling orthogonality}(DO) as a regularization term in the final loss function ensures that the K parallel convolutions in downsampling, along with their distinct channels, are responsible for extracting different features during network training. During inference, these K parallel convolutions, which extract diverse features, can be reparameterized into a single convolution. As a result, \textit{downsampling orthogonality} achieves the aim of encoding diverse features of small objects using only one downsampling convolution during model inference. 
\subsection{Loss Function}
\label{subsec:Loss Function}

The multi-task loss function of the proposed SpirDet predominantly comprises three components: (1) Output SoftIoU Loss $\mathcal{L}_{Output}$, which measures the discrepancy between the final output $O$ of SpirDet and the Ground Truth.  (2) Sparse SoftIoU Loss $\mathcal{L}_{Sparse}$, which quantifies the difference between the coarse-grained probability map $\mathbf{V}$, generated by the \textit{coarse head} of the DBSD, and the coarse-grained Ground Truth. (3) The \textit{orthogonality regularization} loss $\mathcal{L}_{orth}$, which acts as a regularization term within downsampling orthogonality.  Consequently, the aggregate loss function can be expressed as:
\begin{equation}
\mathcal{L}_{all} = \mathcal{L}_{Output} + \mathcal{L}_{Sparse} + \mathcal{L}_{orth}.
\end{equation}


\section{Experiment}
Table \ref{tab:dataset} presents the four publicly accessible infrared small target datasets utilized in our experiments: NUDT-SIRST\cite{DNANET}, IRSTD-1K\cite{isnet}, SIRST3, and NUST\cite{MSvsFA}. SIRST3 is an amalgamated dataset from the NUAA-SIRST\cite{acm}, NUDT-SIRST\cite{DNANET}, and IRSTD-1K\cite{isnet} datasets. For the NUST\cite{MSvsFA} dataset, we randomly divide it into train set(80\%) and test set(20\%).

\begin{table}[h]
\centering
\resizebox{0.9\columnwidth}{!}{%
\begin{tabular}{|l|c|c|c|c|}
\hline
Dataset    & Resolution & Maximum & Average & Train/Test \\ \hline
NUDT-SIRST & 256        & 0.27\%  & 0.066\% & 664/665    \\
IRSTD-1K   & 512        & 0.68\%  & 0.028\% & 801/202   \\
SIRST3     & multiscale & 0.68\%  & 0.052\% & 1677/1080  \\
NUST       & 128        & 1.07\%  & 0.239\% & 7895/1998  \\ \hline
\end{tabular}%
}
\\[0.1cm]
\caption{Attributes of publicly accessible infrared small target datasets: Resolution: image size; Maximum, Average: the maximum and average proportions of small targets within the dataset.  Train/test: the number of images. }
\label{tab:dataset}
\end{table}

\subsection{Implementation Details}
\label{Implementation Details}
All models' training and experimental procedures were executed on an NVIDIA GeForce RTX 3090 GPU. For the IRSTD-1K\cite{isnet}, NUDT-SIRST\cite{DNANET}, SIRST3, and NUST\cite{MSvsFA} datasets. We adhered to a training schedule of 400, 3000, 3000, and 2000 epochs for those datasets, respectively. The AdamW optimizer was utilized, accompanied by Cosine Annealing for learning rate decay, with an initial value of 0.0015 decaying to a minimum of 0.0005. For the multi-scale SIRST3 dataset, all images were uniformly resized to a $256\times256$ resolution for the experiments. We assessed model performance employing the pixel-level metric of Mean Intersection over Union ($MIoU$), object-level metrics of probability of detection ($P_d$) and False-alarm rate ($F_a$), and evaluated model speed in terms of $FPS$.
\begin{table}[h]
\centering
\resizebox{0.6\columnwidth}{!}{%
\begin{tabular}{|cc|cccc|}
\hline
\multicolumn{2}{|c|}{SpirDet} &
  lr &
  t &
  s &
  m \\ \hline
\multicolumn{1}{|c|}{} &
  $512\times512$ &
  \cellcolor[HTML]{EFEFEF}\textbf{-} &
  \cellcolor[HTML]{EFEFEF}\textbf{1} &
  \cellcolor[HTML]{EFEFEF}\textbf{1} &
  \cellcolor[HTML]{EFEFEF}\textbf{2} \\
\multicolumn{1}{|c|}{} &
  $256\times256$ &
  \cellcolor[HTML]{EFEFEF}\textbf{4} &
  \cellcolor[HTML]{EFEFEF}\textbf{2} &
  \cellcolor[HTML]{EFEFEF}\textbf{4} &
  \cellcolor[HTML]{EFEFEF}\textbf{6} \\
\multicolumn{1}{|c|}{} &
  $128\times128$ &
  \cellcolor[HTML]{EFEFEF}\textbf{2} &
  \cellcolor[HTML]{EFEFEF}\textbf{2} &
  \cellcolor[HTML]{EFEFEF}\textbf{4} &
  \cellcolor[HTML]{EFEFEF}\textbf{8} \\
\multicolumn{1}{|c|}{} &
  $64\times64$ &
  \cellcolor[HTML]{EFEFEF}\textbf{2} &
  \cellcolor[HTML]{EFEFEF}\textbf{1} &
  \cellcolor[HTML]{EFEFEF}\textbf{1} &
  \cellcolor[HTML]{EFEFEF}\textbf{1} \\
\multicolumn{1}{|c|}{\multirow{-5}{*}{Num. of Blocks}} &
  $32\times32$ &
  \cellcolor[HTML]{EFEFEF}\textbf{1} &
  \cellcolor[HTML]{EFEFEF}\textbf{-} &
  \cellcolor[HTML]{EFEFEF}\textbf{-} &
  \cellcolor[HTML]{EFEFEF}\textbf{-} \\ \hline
\multicolumn{2}{|c|}{Num. of SparseConvs} &
  \cellcolor[HTML]{EFEFEF}\textbf{2} &
  \cellcolor[HTML]{EFEFEF}\textbf{4} &
  \cellcolor[HTML]{EFEFEF}\textbf{4} &
  \cellcolor[HTML]{EFEFEF}\textbf{4} \\ \hline
\end{tabular}%
}
\\[0.2cm]
\caption{Detail configurations of SpirDet. `Num. of Blocks' denotes the number of blocks within each RepBlock, while `Num. of SparseConvs' signifies the count of sparse convolutions in the SparseHead.}
\label{tab:config}
\end{table}

Table \ref{tab:config} illustrates that our proposed SpirDet incorporates detail configurations to cater to infrared small target detection across varying resolutions. The `lr' configuration is tailored for low resolution, corresponding to the NUDT-SIRST\cite{DNANET}, SIRST3, and NUST\cite{MSvsFA} datasets. The `t', `s', `m' configurations are designed to manage high-resolution data exceeding $512\times512$, such as the IRSTD-1K\cite{isnet} dataset, and respectively symbolize the tiny, small, and medium variants of SpirDet.

\begin{table}[h]
\resizebox{\columnwidth}{!}{%
\begin{tabular}{|ll|l|c|cccc|}
\hline
\multicolumn{2}{|c|}{} &
  \multicolumn{1}{c|}{} &
   &
  \multicolumn{4}{c|}{IRSTD-1K} \\ \cline{5-8} 
\multicolumn{2}{|c|}{\multirow{-2}{*}{Methods}} &
  \multicolumn{1}{c|}{\multirow{-2}{*}{Venue}} &
  \multirow{-2}{*}{Params} &
  \multicolumn{1}{c|}{$MIou\uparrow$} &
  \multicolumn{1}{c|}{$FPS\uparrow$} &
  \multicolumn{1}{c|}{$P_d\uparrow$} &
  $F_a\downarrow$ \\ \hline
\multicolumn{1}{|l|}{} &
  Top-Hat &
  OE1996 &
  - &
  \multicolumn{1}{c|}{10.06} &
  \multicolumn{1}{c|}{-} &
  \multicolumn{1}{c|}{75.11} &
  143.20 \\
\multicolumn{1}{|l|}{} &
  Max-Median &
  SPIE1999 &
  - &
  \multicolumn{1}{c|}{7.003} &
  \multicolumn{1}{c|}{-} &
  \multicolumn{1}{c|}{65.21} &
  5.97 \\
\multicolumn{1}{|l|}{} &
  RLCM &
  GRSL2018 &
  - &
  \multicolumn{1}{c|}{14.62} &
  \multicolumn{1}{c|}{-} &
  \multicolumn{1}{c|}{65.66} &
  1.79 \\
\multicolumn{1}{|l|}{} &
  WSLCM &
  GRSL2021 &
  - &
  \multicolumn{1}{c|}{0.98} &
  \multicolumn{1}{c|}{-} &
  \multicolumn{1}{c|}{70.02} &
  1502.70 \\
\multicolumn{1}{|l|}{} &
  TLLCM &
  GRSL2019 &
  - &
  \multicolumn{1}{c|}{5.36} &
  \multicolumn{1}{c|}{-} &
  \multicolumn{1}{c|}{63.97} &
  0.49 \\
\multicolumn{1}{|l|}{} &
  MSLCM &
  IPT2018 &
  - &
  \multicolumn{1}{c|}{5.34} &
  \multicolumn{1}{c|}{-} &
  \multicolumn{1}{c|}{59.93} &
  0.54 \\
\multicolumn{1}{|l|}{} &
  MSPCM &
  IPT2018 &
  - &
  \multicolumn{1}{c|}{7.33} &
  \multicolumn{1}{c|}{-} &
  \multicolumn{1}{c|}{60.27} &
  1.52 \\
\multicolumn{1}{|l|}{} &
  IPI &
  TIP2013 &
  - &
  \multicolumn{1}{c|}{27.92} &
  \multicolumn{1}{c|}{-} &
  \multicolumn{1}{c|}{81.37} &
  1.61 \\
\multicolumn{1}{|l|}{} &
  NRAM &
  RS2018 &
  - &
  \multicolumn{1}{c|}{15.24} &
  \multicolumn{1}{c|}{-} &
  \multicolumn{1}{c|}{70.67} &
  1.69 \\
\multicolumn{1}{|l|}{} &
  RIPT &
  JSTARS2018 &
  - &
  \multicolumn{1}{c|}{14.10} &
  \multicolumn{1}{c|}{-} &
  \multicolumn{1}{c|}{77.54} &
  2.83 \\
\multicolumn{1}{|l|}{} &
  PSTNN &
  RS2019 &
  - &
  \multicolumn{1}{c|}{24.57} &
  \multicolumn{1}{c|}{-} &
  \multicolumn{1}{c|}{71.98} &
  3.52 \\
\multicolumn{1}{|l|}{\multirow{-12}{*}{1}} &
  MSLSTIPT &
  RS2023 &
  - &
  \multicolumn{1}{c|}{11.43} &
  \multicolumn{1}{c|}{-} &
  \multicolumn{1}{c|}{79.02} &
  152.40 \\ \hline
\multicolumn{1}{|l|}{} &
  ACM &
  WACV2021 &
  0.40 &
  \multicolumn{1}{c|}{59.15} &
  \multicolumn{1}{c|}{162.59} &
  \multicolumn{1}{c|}{90.57} &
  2.04 \\
\multicolumn{1}{|l|}{} &
  ALCNet &
  TGRS2021 &
  0.43 &
  \multicolumn{1}{c|}{61.59} &
  \multicolumn{1}{c|}{{\ul 214.99}} &
  \multicolumn{1}{c|}{89.56} &
  1.44 \\
\multicolumn{1}{|l|}{} &
  ISNet &
  CVPR2022 &
  0.97 &
  \multicolumn{1}{c|}{61.85} &
  \multicolumn{1}{c|}{28.41} &
  \multicolumn{1}{c|}{90.23} &
  3.15 \\
\multicolumn{1}{|l|}{} &
  RDIAN &
  TGRS2023 &
  \textbf{0.22} &
  \multicolumn{1}{c|}{59.93} &
  \multicolumn{1}{c|}{32.74} &
  \multicolumn{1}{c|}{87.20} &
  3.32 \\
\multicolumn{1}{|l|}{} &
  DNA-Net &
  TIP2023 &
  4.70 &
  \multicolumn{1}{c|}{64.88} &
  \multicolumn{1}{c|}{6.65} &
  \multicolumn{1}{c|}{89.22} &
  2.59 \\
\multicolumn{1}{|l|}{} &
  ISTDU-Net &
  GRSL2022 &
  2.75 &
  \multicolumn{1}{c|}{65.71} &
  \multicolumn{1}{c|}{26.58} &
  \multicolumn{1}{c|}{90.57} &
  1.37 \\
  \multicolumn{1}{|l|}{} &
  LWNet &
  TGRS2023 &
  - &
  \multicolumn{1}{c|}{{\ul 67.55}} &
  \multicolumn{1}{c|}{161.00} &
  \multicolumn{1}{c|}{90.23} &
  3.6 \\
  \multicolumn{1}{|l|}{} &
  RepISD &
  TGRS2023 &
  - &
  \multicolumn{1}{c|}{66.38} &
  \multicolumn{1}{c|}{58.17} &
  \multicolumn{1}{c|}{88.55} &
  1.49 \\
\multicolumn{1}{|l|}{} &
  UIU-Net &
  TIP2023 &
  50.54 &
  \multicolumn{1}{c|}{65.69} &
  \multicolumn{1}{c|}{22.49} &
  \multicolumn{1}{c|}{{\ul 91.24}} &
  1.34 \\ \cline{2-8} 
\multicolumn{1}{|l|}{} &
  \cellcolor[HTML]{EFEFEF}\textbf{SpirDet-t} &
  \cellcolor[HTML]{EFEFEF}\textbf{Ours} &
  \cellcolor[HTML]{EFEFEF}{\ul 0.23} &
  \multicolumn{1}{c|}{\cellcolor[HTML]{EFEFEF}64.75} &
  \multicolumn{1}{c|}{\cellcolor[HTML]{EFEFEF}\textbf{258.60}} &
  \multicolumn{1}{c|}{\cellcolor[HTML]{EFEFEF}88.88} &
  \cellcolor[HTML]{EFEFEF}\textbf{0.54} \\
\multicolumn{1}{|l|}{} &
  \cellcolor[HTML]{EFEFEF}\textbf{SpirDet-s} &
  \cellcolor[HTML]{EFEFEF}\textbf{Ours} &
  \cellcolor[HTML]{EFEFEF}0.28 &
  \multicolumn{1}{c|}{\cellcolor[HTML]{EFEFEF}67.46} &
  \multicolumn{1}{c|}{\cellcolor[HTML]{EFEFEF}208.96} &
  \multicolumn{1}{c|}{\cellcolor[HTML]{EFEFEF}88.55} &
  \cellcolor[HTML]{EFEFEF}{\ul 1.21} \\
\multicolumn{1}{|l|}{\multirow{-10}{*}{2}} &
  \cellcolor[HTML]{EFEFEF}\textbf{SpirDet-m} &
  \cellcolor[HTML]{EFEFEF}\textbf{Ours} &
  \cellcolor[HTML]{EFEFEF}0.48 &
  \multicolumn{1}{c|}{\cellcolor[HTML]{EFEFEF}\textbf{70.45}} &
  \multicolumn{1}{c|}{\cellcolor[HTML]{EFEFEF}189.38} &
  \multicolumn{1}{c|}{\cellcolor[HTML]{EFEFEF}\textbf{92.59}} &
  \cellcolor[HTML]{EFEFEF}1.28 \\ \hline
\end{tabular}%
}
\\[0.1cm]
\caption{Our Model $v.s.$ SOTAs: Comparison of \textit{Params($M$)}, $MIoU(\%)$, $FPS$, $P_d(\%)$, and $F_a(\times10^{-5})$ Values on the IRSTD-1K\cite{isnet} Dataset. The best results are highlighted in bold, and the second are in underline. Methods classified under `1' are model-based, while those under `2' are data-driven.}
\label{tab:1K}
\end{table}

\begin{table}[h]
\centering
\resizebox{\columnwidth}{!}{%
\begin{tabular}{|l|l|c|cccc|}
\hline
\multicolumn{1}{|c|}{} &
  \multicolumn{1}{c|}{} &
   &
  \multicolumn{4}{c|}{NUDT-SIRST} \\ \cline{4-7} 
\multicolumn{1}{|c|}{\multirow{-2}{*}{Methods}} &
  \multicolumn{1}{c|}{\multirow{-2}{*}{Venue}} &
  \multirow{-2}{*}{Params} &
  \multicolumn{1}{c|}{$MIoU\uparrow$} &
  \multicolumn{1}{c|}{$FPS\uparrow$} &
  \multicolumn{1}{c|}{$P_d\uparrow$} &
  $F_a\downarrow$ \\ \hline
Top-Hat &
  OE1996 &
  - &
  \multicolumn{1}{c|}{20.72} &
  \multicolumn{1}{c|}{-} &
  \multicolumn{1}{c|}{78.40} &
  16.67 \\
Max-Median &
  SPIE1999 &
  - &
  \multicolumn{1}{c|}{4.20} &
  \multicolumn{1}{c|}{-} &
  \multicolumn{1}{c|}{58.41} &
  3.68 \\
RLCM &
  GRSL2018 &
  - &
  \multicolumn{1}{c|}{15.13} &
  \multicolumn{1}{c|}{-} &
  \multicolumn{1}{c|}{66.34} &
  16.29 \\
WSLCM &
  GRSL2021 &
  - &
  \multicolumn{1}{c|}{0.84} &
  \multicolumn{1}{c|}{-} &
  \multicolumn{1}{c|}{74.57} &
  5239.16 \\
TLLCM &
  GRSL2019 &
  - &
  \multicolumn{1}{c|}{7.05} &
  \multicolumn{1}{c|}{-} &
  \multicolumn{1}{c|}{62.01} &
  4.61 \\
MSLCM &
  IPT2018 &
  - &
  \multicolumn{1}{c|}{6.64} &
  \multicolumn{1}{c|}{-} &
  \multicolumn{1}{c|}{56.82} &
  2.56 \\
MSPCM &
  IPT2018 &
  - &
  \multicolumn{1}{c|}{5.85} &
  \multicolumn{1}{c|}{-} &
  \multicolumn{1}{c|}{55.86} &
  11.59 \\
IPI &
  TIP2013 &
  - &
  \multicolumn{1}{c|}{17.75} &
  \multicolumn{1}{c|}{-} &
  \multicolumn{1}{c|}{74.48} &
  4.12 \\
NRAM &
  RS2018 &
  - &
  \multicolumn{1}{c|}{6.93} &
  \multicolumn{1}{c|}{-} &
  \multicolumn{1}{c|}{56.40} &
  1.92 \\
RIPT &
  JSTARS2018 &
  - &
  \multicolumn{1}{c|}{29.44} &
  \multicolumn{1}{c|}{-} &
  \multicolumn{1}{c|}{91.85} &
  34.43 \\
PSTNN &
  RS2019 &
  - &
  \multicolumn{1}{c|}{14.84} &
  \multicolumn{1}{c|}{-} &
  \multicolumn{1}{c|}{66.13} &
  4.41 \\
MSLSTIPT &
  RS2023 &
  - &
  \multicolumn{1}{c|}{8.34} &
  \multicolumn{1}{c|}{-} &
  \multicolumn{1}{c|}{47.39} &
  8.81 \\ \hline
ACM &
  WACV2021 &
  0.40 &
  \multicolumn{1}{c|}{64.85} &
  \multicolumn{1}{c|}{{\ul 330.64}} &
  \multicolumn{1}{c|}{96.72} &
  2.85 \\
ALCNet &
  TGRS2021 &
  0.43 &
  \multicolumn{1}{c|}{61.13} &
  \multicolumn{1}{c|}{\textbf{346.58}} &
  \multicolumn{1}{c|}{97.24} &
  2.90 \\
ISNet &
  CVPR2022 &
  0.97 &
  \multicolumn{1}{c|}{81.23} &
  \multicolumn{1}{c|}{80.08} &
  \multicolumn{1}{c|}{97.77} &
  0.63 \\
RDIAN &
  TGRS2023 &
  \textbf{0.22} &
  \multicolumn{1}{c|}{82.41} &
  \multicolumn{1}{c|}{121.89} &
  \multicolumn{1}{c|}{98.83} &
  1.36 \\
DNA-Net &
  TIP2023 &
  4.70 &
  \multicolumn{1}{c|}{89.81} &
  \multicolumn{1}{c|}{31.35} &
  \multicolumn{1}{c|}{{\ul 98.90}} &
  0.64 \\
ISTDU-Net &
  GRSL2022 &
  2.75 &
  \multicolumn{1}{c|}{{\ul 92.34}} &
  \multicolumn{1}{c|}{98.98} &
  \multicolumn{1}{c|}{98.51} &
  {\ul 0.55} \\
LWNet &
  TGRS2023 &
  - &
  \multicolumn{1}{c|}{73.91} &
  \multicolumn{1}{c|}{183.00} &
  \multicolumn{1}{c|}{95.13} &
   2.76 \\
RepISD &
  TGRS2023 &
  - &
  \multicolumn{1}{c|}{-} &
  \multicolumn{1}{c|}{-} &
  \multicolumn{1}{c|}{-} &
   - \\
UIU-Net &
  TIP2023 &
  50.54 &
  \multicolumn{1}{c|}{90.51} &
  \multicolumn{1}{c|}{54.17} &
  \multicolumn{1}{c|}{98.83} &
  0.83 \\ \hline
\rowcolor[HTML]{EFEFEF} 
\textbf{SpirDet-lr} &
  \textbf{Ours} &
  {\ul 0.24} &
  \multicolumn{1}{c|}{\cellcolor[HTML]{EFEFEF}\textbf{94.43}} &
  \multicolumn{1}{c|}{\cellcolor[HTML]{EFEFEF}310.15} &
  \multicolumn{1}{c|}{\cellcolor[HTML]{EFEFEF}\textbf{99.15}} &
  \textbf{0.54} \\ \hline
\end{tabular}%
}
\\[0.1cm]
\caption{Our Model $v.s.$ SOTAs: Comparison of \textit{Params($M$)}, $MIoU(\%)$, $FPS$, $P_d(\%)$, and $F_a(\times10^{-5})$ Values on the NUDT-SIRST\cite{DNANET} Dataset.}
\label{tab:NUDT}
\end{table}
\begin{table*}[h]
\centering
\resizebox{\textwidth}{!}{%
\begin{tabular}{|l|l|c|cccc|cccc|cccc|}
\hline
\multicolumn{1}{|c|}{} &
  \multicolumn{1}{c|}{} &
   &
  \multicolumn{4}{c|}{SIRST3} &
  \multicolumn{4}{c|}{NUST} &
  \multicolumn{4}{c|}{Average} \\ \cline{4-15} 
\multicolumn{1}{|c|}{\multirow{-2}{*}{Methods}} &
  \multicolumn{1}{c|}{\multirow{-2}{*}{Venue}} &
  \multirow{-2}{*}{Params} &
  \textit{$MIoU\uparrow$} &
  \textit{$FPS\uparrow$} &
  \textit{$P_d\uparrow$} &
  \textit{$F_a\downarrow$} &
  \textit{$MIoU\uparrow$} &
  \textit{$FPS\uparrow$} &
  \textit{$P_d\uparrow$} &
  \textit{$F_a\downarrow$} &
  \textit{$MIoU\uparrow$} &
  \textit{$FPS\uparrow$} &
  \textit{$P_d\uparrow$} &
  \textit{$F_a\downarrow$} \\ \hline

ACM &
  WACV2021 &
  0.40 &
  57.98 &
  {\ul 272.15} &
  86.67 &
  1.29 &
  79.95 &
  294.67 &
  91.22 &
  0.63 &
  68.97 &
  283.41 &
  88.95 &
  0.96 \\
ALCNet &
  TGRS2021 &
  0.43 &
  66.03 &
  \textbf{303.23} &
  93.30 &
  1.13 &
  86.20 &
  {\ul 317.39} &
  95.46 &
  1.08 &
  76.12 &
  \textbf{310.31} &
  94.38 &
  1.11 \\
ISNet &
  CVPR2022 &
  0.97 &
  71.83 &
  77.17 &
  95.06 &
  1.93 &
  90.10 &
  114.24 &
  96.27 &
  \textbf{0.22} &
  80.97 &
  95.71 &
  95.67 &
  1.08 \\
RDIAN &
  TGRS2023 &
  \textbf{0.22} &
  70.97 &
  118.80 &
  93.50 &
  1.86 &
  90.13 &
  310.29 &
  96.24 &
  0.47 &
  80.55 &
  214.55 &
  94.87 &
  1.17 \\
DNA-Net &
  TIP2023 &
  4.70 &
  75.67 &
  30.57 &
  {\ul 96.14} &
  1.27 &
  88.79 &
  71.22 &
  96.68 &
  0.73 &
  82.23 &
  50.90 &
  96.41 &
  1.00 \\
ISTDU-Net &
  GRSL2022 &
  2.75 &
  {\ul 79.18} &
  98.76 &
  96.07 &
  \textbf{0.70} &
  91.05 &
  112.29 &
  97.41 &
  0.39 &
  {\ul 85.12} &
  105.53 &
  {\ul 96.74} &
  {\ul 0.55} \\
LWNet &
  TGRS2023 &
  - &
  67.39 &
  - &
  92.95 &
  \textbf{5.56} &
  86.07 &
  - &
   94.73 &
  3.03 &
  76.73 &
  - &
  93.84 &
  4.30 \\
RepISD &
  TGRS2023 &
  - &
  - &
  - &
  - &
  \textbf{-} &
  \textbf{92.47} &
  - &
 \textbf{97.56} &
  {\ul 0.28} &
  - &
  - &
  - &
  - \\
UIU-Net &
  TIP2023 &
  50.54 &
  76.89 &
  52.51 &
  96.00 &
  0.93 &
  90.23 &
  56.97 &
  96.68 &
  0.30 &
  83.56 &
  54.74 &
  96.34 &
  0.62 \\
\rowcolor[HTML]{EFEFEF} 
\textbf{SpirDet-lr} &
  \textbf{Ours} &
  {\ul 0.28} &
  \textbf{82.45} &
  250.42 &
  \textbf{96.54} &
  {\ul 0.78} &
  {\ul 91.15} &
  \textbf{369.05} &
  {\ul 97.41} &
   0.32 &
  \textbf{86.80} &
  {\ul 309.74} &
  \textbf{96.98} &
  \textbf{0.55} \\ \hline
\end{tabular}%
}
\\[0.1cm]
\caption{Our Model $v.s.$ SOTAs: Comparison of \textit{Params($M$)}, $MIoU(\%)$, $FPS$, $P_d(\%)$, and $F_a(\times10^{-5})$ Values on the SIRST3 and NUST\cite{MSvsFA} Datasets.}
\label{tab:nust}
\end{table*}

\subsection{Effectiveness of Our Approach}
\label{Effectiveness of Our Approach}
A comprehensive comparison was conducted between the proposed SpirDet and state-of-the-art methods, encompassing Top-Hat\cite{tophat}, Max-Median\cite{maxmedian}, RLCM\cite{rlcm}, WSLCM\cite{wslcm}, TLLCM\cite{tllcm}, MSLCM\cite{MSLCM}, MSPCM\cite{MSLCM}, IPI\cite{ipi}, NRAM\cite{nram}, RIPT\cite{RIPT}, PSTNN\cite{PSTNN}, MSLSTIPT\cite{MSLSTIPT}, ACM\cite{acm}, ALCNet\cite{ALCNet}, ISNet\cite{isnet}, RDIAN\cite{rdian}, DNANet\cite{DNANET}, ISTDUNet\cite{istdunet}, LWNet\cite{lwnet}, RepISD\cite{repisd}, UIUNet\cite{uiunet}.

\begin{figure}[h]
\centering
\centerline{\includegraphics[width=0.42\textwidth,trim=10 0 0 10,clip]{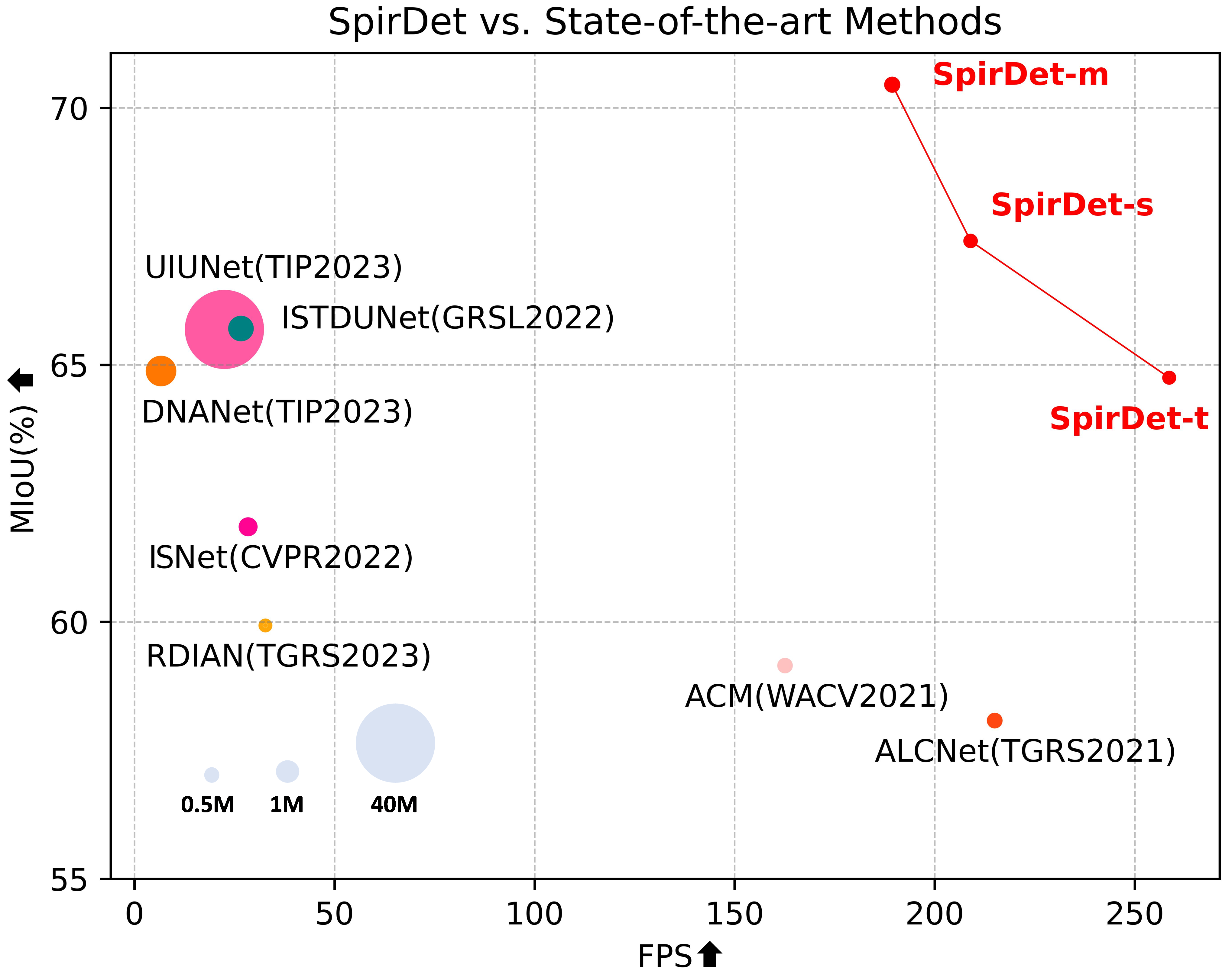}}
\caption{The $MIoU$, $FPS$ versus model size on IRSTD-1K\cite{isnet} test set. Our SpirDet are marked as red point.}\label{fig:score2}
\end{figure}

In Table \ref{tab:1K}, various models were compared on the IRSTD-1K\cite{isnet} dataset, characterized by the smallest target proportion and highest resolution. Notably, our SpirDet-m recorded a 70.45 $MIoU$, exceeding the previous best of 65.71 (ISTDU-Net) by approximately 4.74, with a 2-unit increase in $P_a$, a lower $F_a$, and an approximately sevenfold $FPS$ advantage. And the model parameters were a mere 17\% of ISTDU-Net. The lighter SpirDet-t outperformed other lightweight networks in $MIoU$, reaching 258 $FPS$, with the lowest $F_a$ among all models. In most indicators, our method emerged superior, boasting lower parameters and higher $FPS$. It verifies that our method's significant ability in high-resolution scenarios as shown in Fig.\ref{fig:score2}.

On the distinguishable NUDT-SIRST\cite{DNANET} dataset, as illustrated in Table \ref{tab:NUDT}, SpirDet was tested against the SOTA model. Our method yielded a 94.43 $MIoU$, 99.15 $P_d$, and 0.54 $F_a$, successively surpassing the previous best results by 2.09, 0.25, and 0.01, while maintaining an $FPS$ nearly ten times higher than DNANet with the previous highest $P_d$.

In Table \ref{tab:nust}, comparative experiments were performed between SpirDet and other deep learning-based networks on the SIRST3 and NUST\cite{MSvsFA} datasets. The trials demonstrated that our SpirDet nearly topped all performance indicators, all while maintaining excellent speed performance and minimal parameters.

\begin{figure}[h]
\centering
\centerline{\includegraphics[width=0.5\textwidth,trim=20 40 20 45,clip]{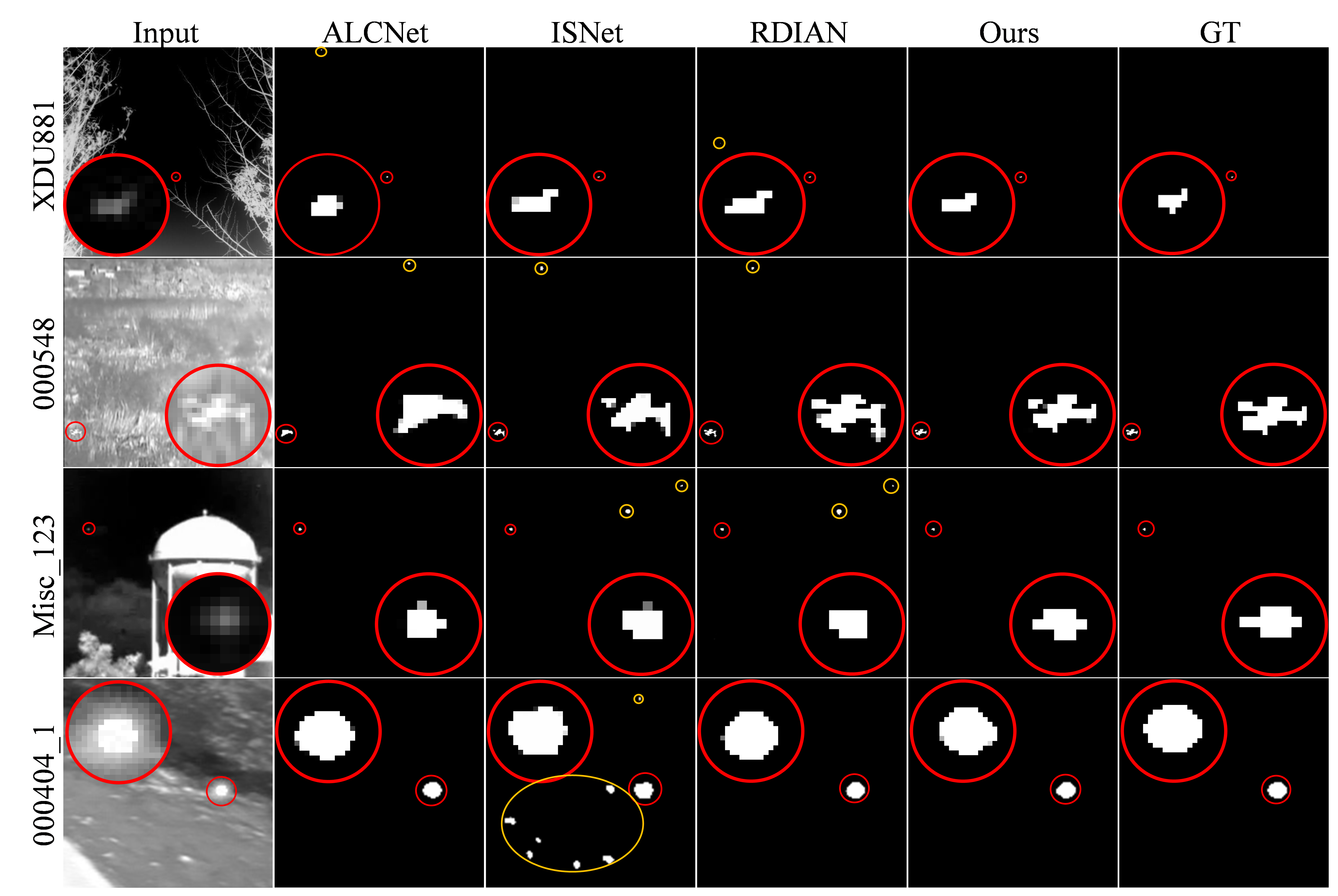}}
\caption{Comparative Visualization of Various Networks on IRSTD-1K\cite{isnet}, NUDT-SIRST\cite{DNANET}, NUAA-SIRST\cite{acm} and  NUST\cite{MSvsFA}  Datasets. The correct detected targets are highlighted in red and magnified, while misidentified targets are marked in yellow.}
\label{fig:compare}
\end{figure}

The preceding results emphatically demonstrate SpirDet’s superior model positioning ability ($P_d$ and $F_a$) and detailed learning capacity ($MIoU$) in complex infrared scenes. Notably, even without resorting to additional acceleration techniques, our model's $FPS$ can match or surpass prior lightweight networks. In addition, we also provide  visual results on four datasets as shown in Fig.\ref{fig:compare}, our proposed SpirDet not only avoids false detections, but also closely aligns the predicted small targets with the Ground Truth.

\subsection{Ablation Studies}
\label{Ablation Studies}
\begin{figure}[h]
\centering
\centerline{\includegraphics[width=0.45\textwidth,trim=100 50 100 50,clip]{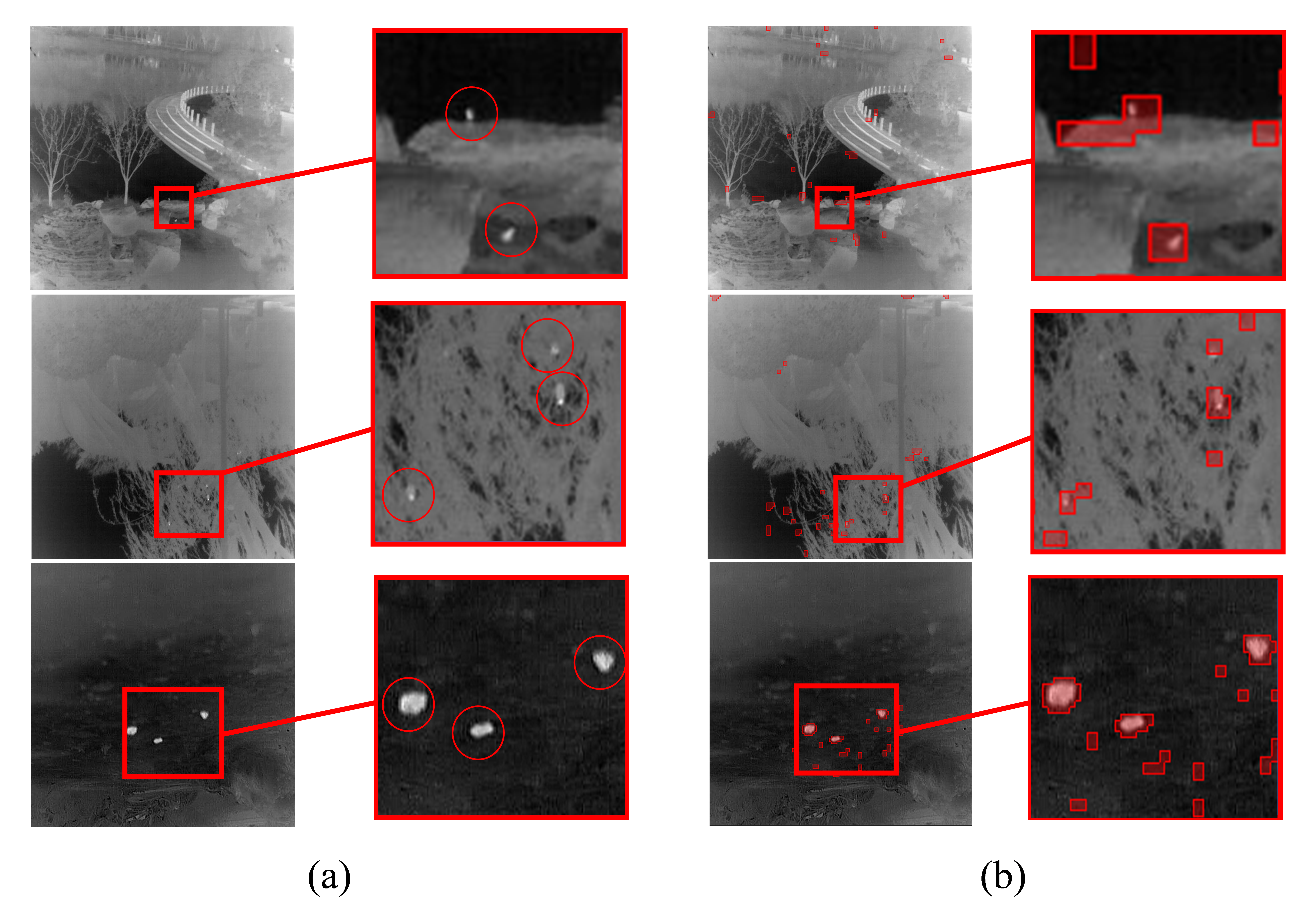}}
\caption{Visualization results of the DBSD. (a)The original images with target locations denoted by red circles. (b)The DBSD's sparse graph, highlighting potential target locations with red zones.}
\label{fig:redzone}
\end{figure}

\begin{table}[h]
\centering

\resizebox{\columnwidth}{!}{%
\begin{tabular}{|l|ccccc|c|}
\hline
\multicolumn{1}{|c|}{\multirow{2}{*}{Metric}} &
  \multicolumn{5}{c|}{w/ DBSD} &
  \multirow{2}{*}{w/o DBSD} \\ \cline{2-6}
\multicolumn{1}{|c|}{} &
  $\alpha$=0.05\% &
  $\alpha$=0.1\% &
  $\alpha$=0.5\% &
  $\alpha$=1\% &
  $\alpha$=5\% &
   \\ \hline
\textit{$MIoU$} &
  38.63 &
  49.56 &
  \textbf{70.45} &
  69.39 &
  68.64 &
  67.17 \\
\textit{$FPS$} & \textbf{204.29} & 191.30 & 189.38 & 178.76 & 173.79 & 159.64 \\
$P_d$ &
  75.42 &
  80.80 &
  \textbf{92.59} &
  91.24 &
  \textit{89.22} &
  89.56 \\
\textit{$F_a$} &
  \textbf{0.34} &
  1.218 &
  1.28 &
  2.226 &
  3.28 &
  3.16 \\ \hline
\end{tabular}%
}
\\[0.1cm]
\caption{The influence of the hyperparameter $\alpha$ in DBSD on the experimental results over the IRSTD-1K\cite{isnet} dataset.}
\label{tab:ab_DBSD}
\end{table}
\noindent
\textbf{The impact of the sparsity ratio $\alpha$ in DBSD.} Table \ref{tab:ab_DBSD} demonstrates that $\alpha$ profoundly affects the outcomes, creating an almost monotonic variation in $FPS$, $P_d$, and $F_a$ in response to changes in $\alpha$. When $\alpha$ is reduced to 0.05\%, it stringently confines the potential area of small targets. As expected, this results in a decrease in the false detection rate, $F_a$, of small targets, and an increase in $FPS$ by curtailing the range the sparse head is required to caculate. Nevertheless, this leads to a decrease in the recall rate for small targets, causing $MIoU$ and $P_d$ to sharply decline to 38.63 and 75.42, respectively. On the contrary, an increase in $\alpha$ causes $F_a$ to rise, but simultaneously leads to a swift decrease in the model's $FPS$, especially when DBSD is not employed, bringing $FPS$ down to approximately 159. After $\alpha$ increases to 0.5\%, $MIoU$ and $P_d$ reach a peak and then start to decline. This leads us to conjecture that $\alpha$ regulates the extent of contextual information for small targets, and an overexpansion of this extent may disrupt the model's learning of small target information. Additionally, we present the visualization results of the sparse graph to demonstrate the effectiveness of DBSD, as shown in Fig. \ref{fig:redzone}.

\begin{table}[h]
\resizebox{\columnwidth}{!}{%
\begin{tabular}{|c|ccc|ccc|}
\hline
                          & \multicolumn{3}{c|}{Params(M)}                        & \multicolumn{3}{c|}{GMACs}                            \\ \cline{2-7} 
\multirow{-2}{*}{Methods} & Before & After & Enhancement                          & Before & After & Enhancement                          \\ \hline
t                         & 0.564  & 0.234 & \cellcolor[HTML]{EFEFEF}\textbf{2.4$\times$} & 1.862  & 0.617 & \cellcolor[HTML]{EFEFEF}\textbf{3.0$\times$} \\
s                         & 0.756  & 0.284 & \cellcolor[HTML]{EFEFEF}\textbf{2.6$\times$} & 2.310  & 0.807 & \cellcolor[HTML]{EFEFEF}\textbf{2.9$\times$} \\
m                         & 1.210  & 0.483 & \cellcolor[HTML]{EFEFEF}\textbf{2.5$\times$} & 2.688  & 1.210 & \cellcolor[HTML]{EFEFEF}\textbf{2.2$\times$} \\ \hline
\end{tabular}%
}
\\[0.1cm]
\caption{Effects of Reparameterization on Params(M) and GMACs in RepirDet.}
\label{tab:repEnhance}
\end{table}
\noindent
\textbf{The ablation studies of the DO-RepEncoder.} In the Table \ref{tab:repEnhance}, we evaluated the enhancements in GMACs and Params that reparameterization brings to SpirDet when applied via DO-RepEncoder. The empirical data indicates that reparameterization, on average, reduces the size of various models by a factor of 2.5. In terms of the GMACs metric the model demonstrates resource efficiency by saving more than twice the computational resources during the inference process. A separate ablation study on the downsampling orthogonality constraint within DO-RepEncoder is presented in Table \ref{tab:DO}. The findings suggest that the  downsampling orthogonality enhances $MIoU$ by 0.5 and $P_d$ by approximately 4.  We typically configure DO-RepEncoder with a parallel convolution count, K, of 4, where the experiments confirm that optimal performance is attained when K is set to 4.

\begin{table}[h]
\centering
\resizebox{0.7\columnwidth}{!}{%
\begin{tabular}{|l|lll|}
\hline
Methods           & $MIoU$        & $P_d$         & $F_a$        \\ \hline
RepEncoder w/ DO  & \textbf{70.4} & \textbf{92.6} & 1.3          \\
RepEncoder w/o DO & 69.9          & 88.8          & \textbf{1.2} \\ \hline
\end{tabular}%
}
\\[0.1cm]
\caption{The impact of Downsampling Orthogonality (DO) on the experimental results on the IRSTD-1K\cite{isnet} dataset.}
\label{tab:DO}
\end{table}

\subsection{Discussions}
\label{Discussions}
\begin{table}[h]
\centering

\resizebox{0.75\columnwidth}{!}{%
\begin{tabular}{|l|l|l|}
\hline
Operator         & Attention         & Sparse        \\ \hline
Context Learning & Self-adaption     & Target-aware  \\
Range            & Global/Local      & Local         \\
Computation      & Soft              & Hard          \\
Purpose          & Relation Modeling & Target Kernel \\ \hline
\end{tabular}%
}
\\[0.1cm]
\caption{The similarities and differences of the attention mechanism and sparse operation in small target detection.}
\label{tab:Atten}
\end{table}

\begin{table}[h]
\centering

\resizebox{0.75\columnwidth}{!}{%
\begin{tabular}{|cc|cccc|}
\hline
                Dataset          &     Mechanism      & $MIoU$        & $FPS$         & $P_d$         & $F_a$        \\ \hline
\multirow{2}{*}{SIRST-3}  & Attention & 74.5          & 30.0          & 95.9          & 1.4          \\
                          & Sparse    & \textbf{75.6} & \textbf{53.3} & \textbf{96.6} & \textbf{0.4} \\ \hline
\multirow{2}{*}{IRSTD-1K} & Attention & \textbf{64.8} & 6.6           & \textbf{89.2} & 2.6          \\
                          & Sparse    & 64.1          & \textbf{18.3} & \textbf{89.2} & \textbf{1.3} \\ \hline
\end{tabular}%
}
\\[0.1cm]
\caption{The impact of the Attention mechanism and Sparse operation on DNA-Net on SIRST-3 and IRSTD-1K\cite{isnet}.}
\label{tab:atten_exp}
\end{table}

\noindent
\textbf{Can the sparse operation replace the attention mechanism?} The attention mechanism\cite{vaswani2017attention,DNANET} and sparse operations, commonly employed in infrared small target detection, possess both similarities and differences, as delineated in Table \ref{tab:Atten}. As outlined in Sec. \ref{Ablation Studies}, the acquisition of excessive contextual information may invite interference. Theoretically, sparse operations can facilitate learning of local contextual information related to the target, while focusing computations solely on potential target positions, thereby accelerating calculations. Thus, it may be a preferable alternative to the attention mechanism. To validate this, we substituted all the attention mechanisms in DNANet with sparse operations, and conducted experiments on the IRSTD-1K and SIRST-3 datasets as shown in Table \ref{tab:atten_exp}. The experimental outcomes reveal that the sparse version of DNANet either matches or surpasses the performance of the attention-based model, while notably enhancing inference speed. 

\begin{figure}[h]
\centering
\centerline{\includegraphics[width=0.34\textwidth]{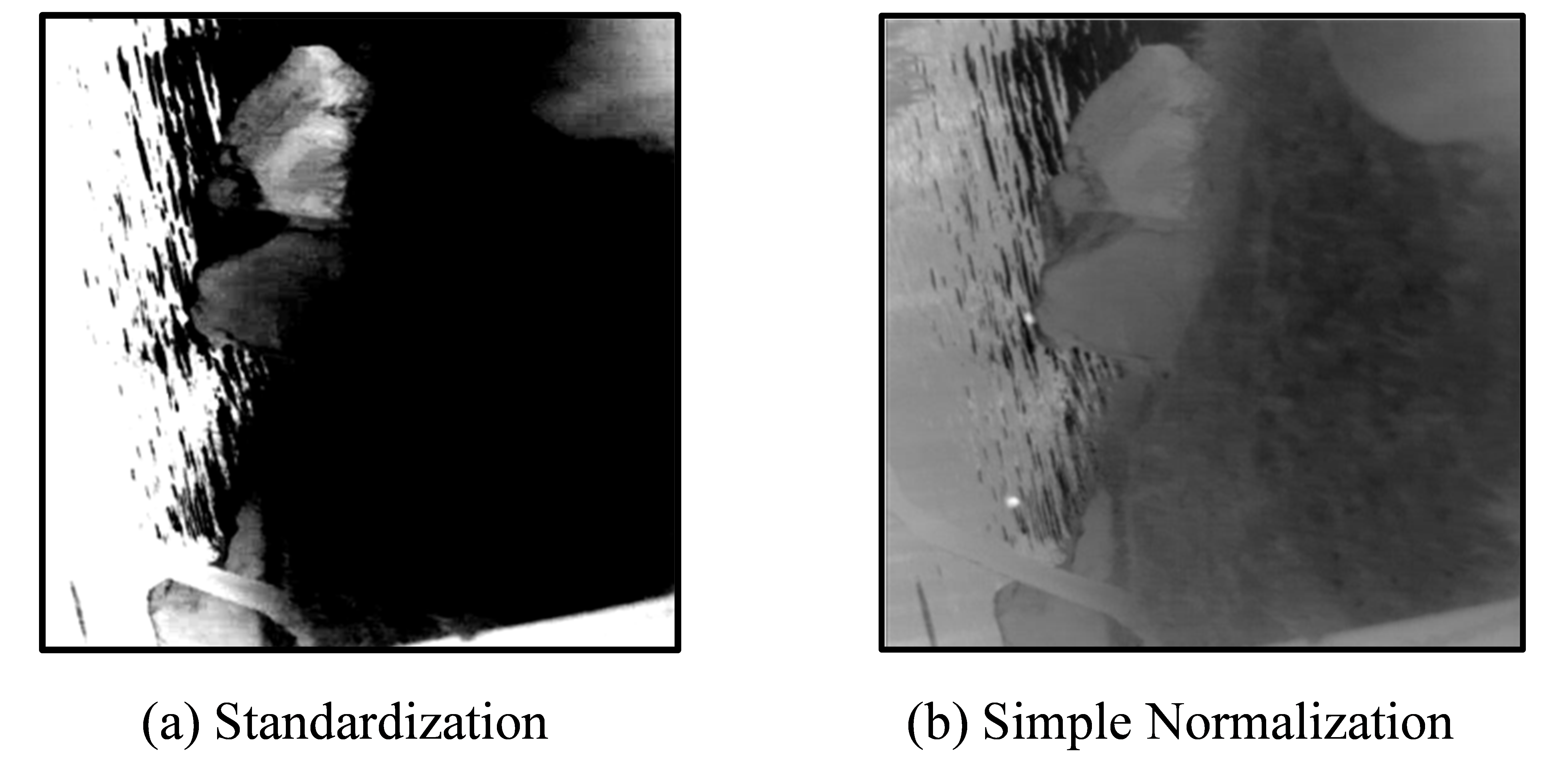}}
\caption{Different normalization methods for input images.}
\label{fig:Normalize}
\end{figure}

\begin{figure*}[h]
\centering
\includegraphics[width=\textwidth,trim=10 0 0 10,clip]{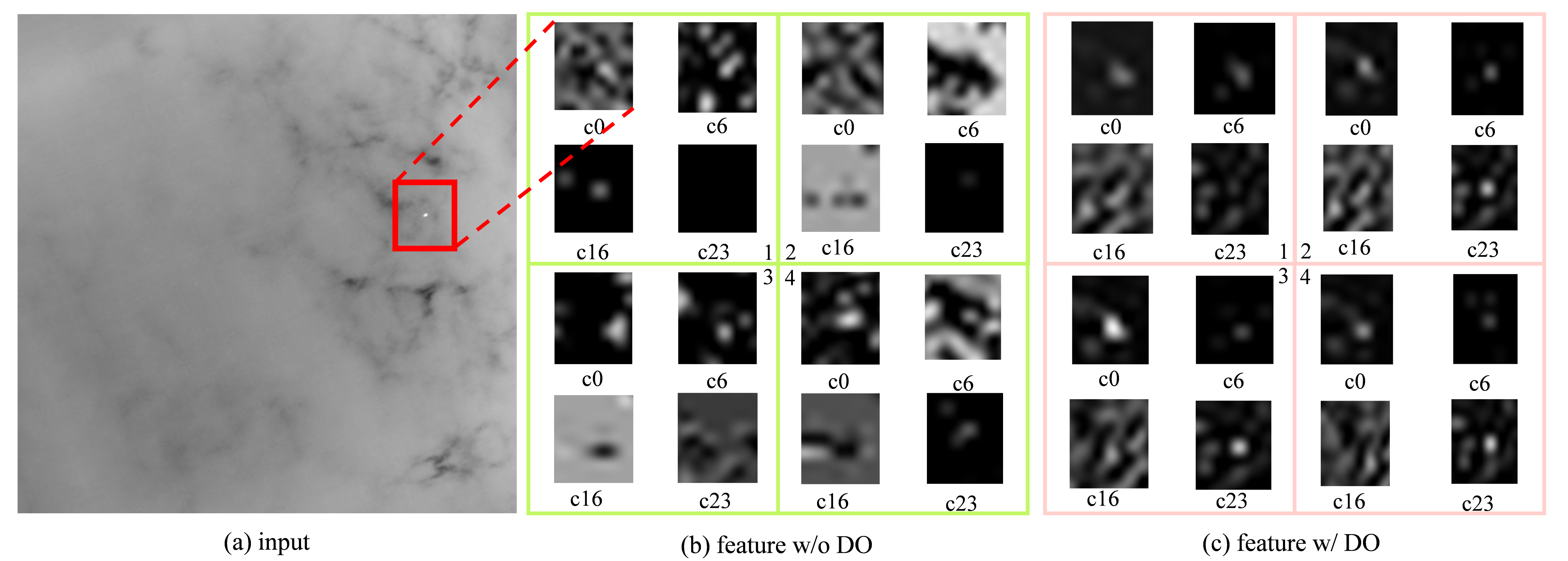}
\caption{Visualization of features around the target for K (K=4) parallel convolutions. (a) Target features when DO was not used. (b) Target features when using DO. `1', `2', `3', `4' represent different convolutions. `c0' represents the $0_{th}$ channel.}
\label{fig:do}
\end{figure*}

\noindent
\textbf{Effects of different Normalizations.} Typically, input images must be mapped to the range [0, 1]. Some infrared small target detectors independently computed the mean and variance of pixel values across different datasets as shown in Fig. \ref{fig:Normalize}(a), and normalized the input images using standardization. This strategy tends to homogenize the pixel value distributions across diverse images. However, standardization may result in diminished image hierarchy. Notably, when the small target and the background exhibit minimal differences, the standardized small target risks being submerged in the background. Experimental results from Table \ref{tab:Normalization} indicate that a straightforward normalization approach, which divides the image by 255 easily as shown in Fig. \ref{fig:Normalize}(b), can more effectively maintain image hierarchy, thereby enhancing the distinguishability of small targets. Thus, we advocate for the use of simple normalization in replace of standardization to normalize input images.
\begin{table}[h]
\centering
\resizebox{0.85\columnwidth}{!}{%
\begin{tabular}{|cc|ccc|}
\hline
\multicolumn{2}{|c|}{Methods}                    & $MIoU$                                & $P_d$                                 & $F_a$                                \\ \hline
                            & w/ Standardization & 68.2                                  & 88.5                                  & 2.0                                  \\
\multirow{-2}{*}{SpirDet-m} & w/ SN              & \cellcolor[HTML]{EFEFEF}\textbf{70.4} & \cellcolor[HTML]{EFEFEF}\textbf{92.5} & \cellcolor[HTML]{EFEFEF}\textbf{1.2} \\ \hline
\end{tabular}%
}
\\[0.1cm]
\caption{The  results of the two normalization methods on the IRSTD-1K\cite{isnet} dataset. `SN' denotes  simple normalization.}
\label{tab:Normalization}
\end{table}

\noindent
\textbf{Analysis of Downsampling Orthogonality. }Fig. \ref{fig:do} visualizes the disparate features surrounding the target, extracted from the feature map $X_2$ after it has been processed by K (K=4) parallel downsampling convolutions. The visualization outcomes highlight two key observations. Firstly, the target features are notably distinct in the  feature maps by the application of Downsampling Orthogonality (DO), whereas the feature maps (without utilizing DO) contain a greater extent of background clutter. Secondly, in the features w/o DO, some channels exhibit a degradation of small target information to all-zero features, a situation not encountered when DO is implemented. In summary, the application of DO to K parallel downsampled convolutions not only effectively inhibits the disappearance of small-target features but also facilitates a more diverse encoding of these features. This enhances the model's capacity to learn features such as the shape of the small targets.

\begin{table}[htbp]
\centering
\resizebox{0.6\columnwidth}{!}{%
\begin{tabular}{|cc|ccc|}
\hline
\multicolumn{2}{|c|}{Feature}                         & $MIoU$           & $P_d$             & $F_a$           \\ \hline
\multicolumn{1}{|c|}{\multirow{3}{*}{SpirDet-m}} & P5 & 68.1           & 87.5           & \textbf{0.9} \\
\multicolumn{1}{|c|}{}                           & P4 & 67.7           & 80.4           & 2.7          \\
\multicolumn{1}{|c|}{}                           & P3 & \textbf{70.4} & \textbf{92.6} & 1.3        \\ \hline
\end{tabular}%
}
\caption{Comparison of $MIoU(\%)$, $P_d(\%)$, and $F_a(\times10^{-5})$ with different layers serving as inputs to the fast branch.}
\label{tab:Fast Branch}
\end{table}

\begin{figure}[h]
\centering
\includegraphics[width=0.48\textwidth,trim=0 370 465 50,clip]{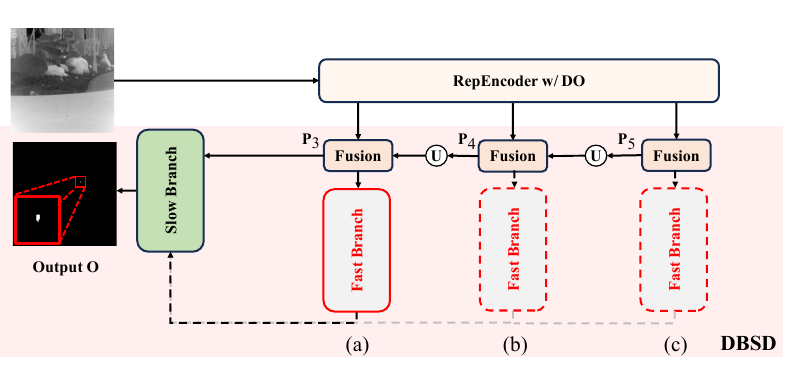}
\caption{Diagram of selecting different layers as input to the fast branch.}
\label{fig:p3p4p5}
\end{figure}

\noindent
\textbf{Which Layer is the Better Input of the Fast Branch? }The role of the fast branch is to provide target potential locations for the refined refinement computation in the slow branch. Therefore, different scales of feature maps as inputs to the fast branch will result in different granularity of potential locations as shown in Fig. \ref{fig:p3p4p5}. In addition, Table \ref{tab:Fast Branch} illustrates the effect of using different resolution feature maps as input to the fast branch. The granularity of potential locations varies with the resolution of the feature maps. Experimental results suggest that maximizing the resolution size of the fast branch, while maintaining a lower resolution, significantly improves outcomes.

\begin{figure*}[htbp]
\centering
\includegraphics[width=\textwidth]{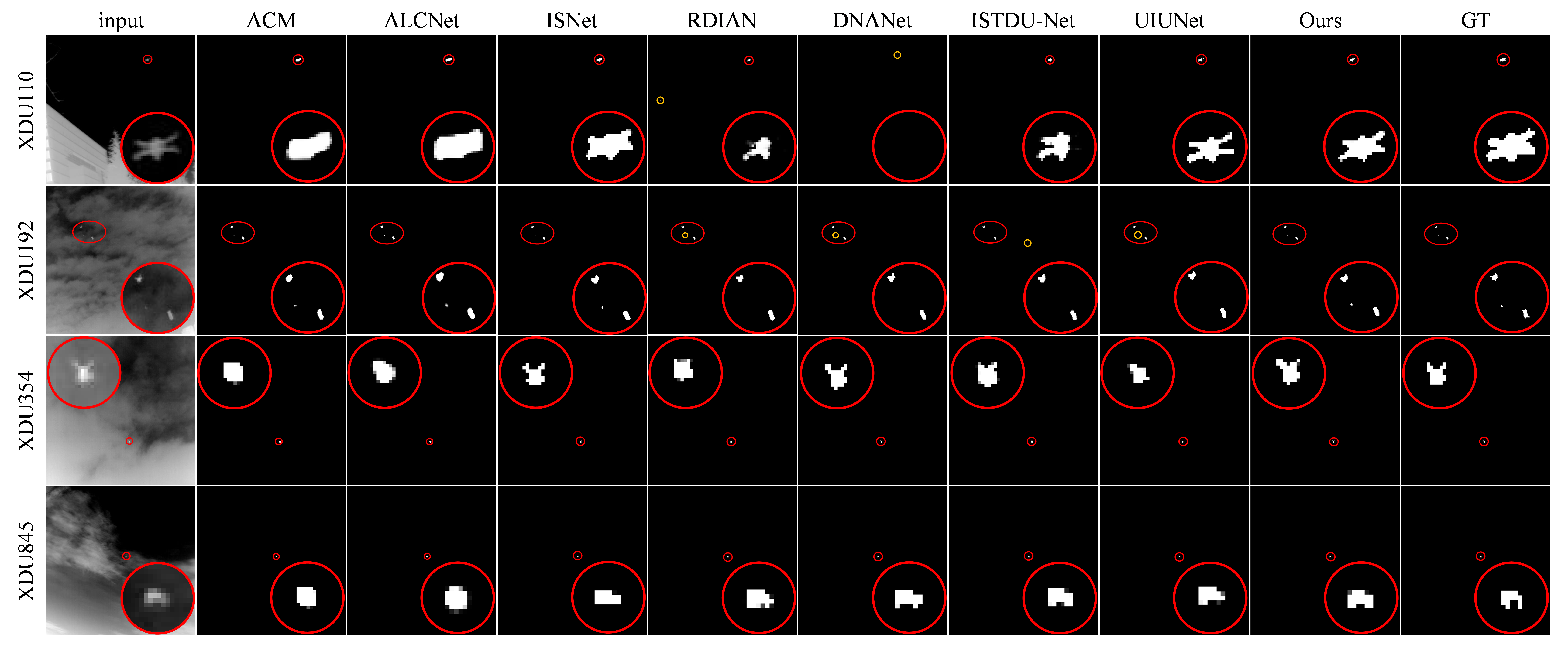}
\caption{Comparative visualization results covering ACM\cite{acm}, ALCNet\cite{ALCNet}, ISNet\cite{isnet}, RDIAN\cite{rdian}, DNANet\cite{DNANET}, ISTDU-Net\cite{istdunet}, UIUNet\cite{uiunet}  on IRSTD-1K\cite{isnet} dataset. The correct detected targets are highlighted in red and magnified, while the misidentified targets are marked in yellow.}
\label{fig:morevis}
\end{figure*}

\begin{figure*}[htbp]
\centering
\includegraphics[width=0.75\textwidth]{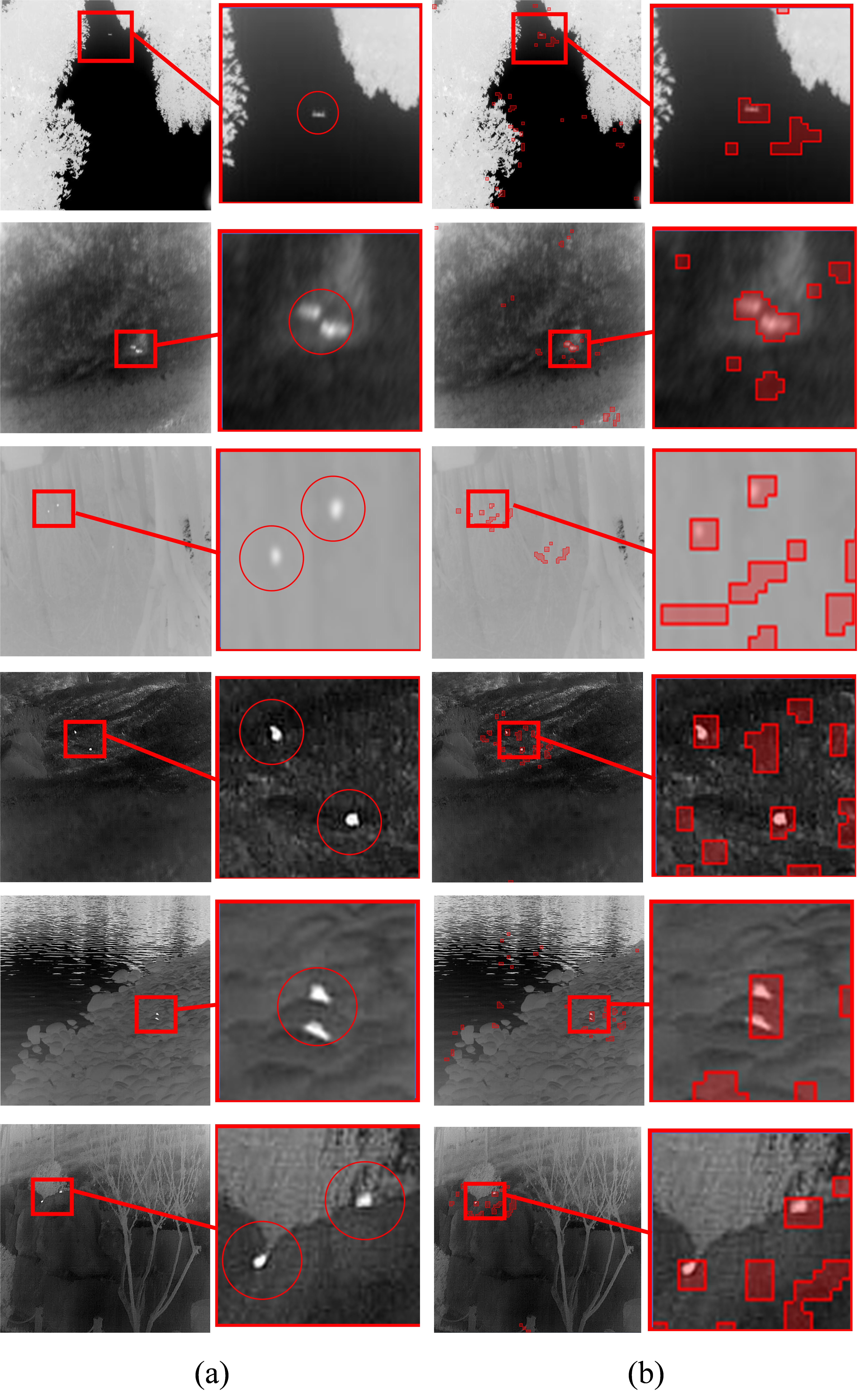}
\caption{More visualization results of the proposed DBSD. (a)The original images with target locations denoted by red circles. (b)The DBSD's sparse graph, highlighting potential target locations with red zones.}
\label{fig:moresparse}
\end{figure*}

\noindent
\textbf{More Visualization Results. }Fig. \ref{fig:morevis} presents the visual results of the proposed method and the SOTA approaches on the  IRSTD-1K\cite{isnet} dataset. As illustrated, SpirDet has a much lower false detection rate, while learning a better shape of small targets. For instance, in  XDU110, XDU354, and XDU845, although many networks are capable of recognizing the presence of small targets, our network exhibits a closer approximation to the Ground Truth (GT) in terms of shape fidelity. For example, the XDU192 image, our method can locate all the small targets whileas the compared methods exhibit omissions. 
The sparse potential target location regions obtained by DBSD are shown in Fig. \ref{fig:moresparse}. All the targets are truly identified by DBSD. Meanwhile,  many possible target location regions are likewise used as candidate regions, which guarantees the recall rate of SpirDet.

\section{Conclusion}
\label{Conclusion}

In this paper, we introduce SpirDet, a novel approach tailored for infrared small target detection. Leveraging sparsity and a reparameterization mechanism, SpirDet efficiently detect small targets on high-resolution feature maps at a lower computational cost. Experimental results on multiple public datasets show that the SpirDet significantly enhancing inference speed while maintaining performance on the four public datasets . In the future, the sparsity and reparameterization mechanisms are expected to be applied in video sequences for detecting infrared small targets.

\bibliographystyle{IEEEtran}
\bibliography{bare_jrnl_new_sample4}


\newpage

 
\vspace{11pt}

\vfill

\end{document}